\documentclass[journal]{IEEEtran}

\usepackage{amsmath,amssymb,amsfonts}
\usepackage{amsthm}
\newtheorem{theorem}{Theorem}
\usepackage{algorithmic}
\usepackage{algorithm}
\usepackage{graphicx}
\usepackage{textcomp}
\usepackage{booktabs}
\usepackage{multirow}
\usepackage{cite}
\usepackage{hyperref}
\usepackage{xcolor}
\usepackage{url}
\usepackage{bm}
\usepackage{mathtools}

\def\BibTeX{{\rm B\kern-.05em{\sc i\kern-.025em b}\kern-.08em
    T\kern-.1667em\lower.7ex\hbox{E}\kern-.125emX}}

\begin{document}

\title{Differential Spectral Damping: Gap-Adaptive Regularization for Ill-Conditioned Kernel Methods}

\author{Praveg~Vashishtha
\thanks{P. Vashishtha is with the Department of Computer Science and Engineering, Indian Institute of Technology Patna, India, and with Sopra Banking Software India as Senior Software Development Engineer. (e-mail: praveg\_pa2503mth259@iitp.ac.in; omvashishtha432@gmail.com). LinkedIn: \url{https://www.linkedin.com/in/pravegvashishtha/}}}

\maketitle

\begin{abstract}
Kernel methods requiring matrix inversion---particularly Least-Squares Twin Support Vector Machines (LSTSVM)---suffer from exponential eigenvalue decay in their system matrices, producing severely ill-conditioned problems where standard Tikhonov regularization applies uniform damping regardless of eigenvector reliability. We propose Differential Spectral Damping (DSD), a regularization formula that adapts its penalty to localized eigengap structure: preserving eigenvectors with large spectral gaps (reliable per Davis-Kahan perturbation theory) while aggressively suppressing those with small gaps (directionally corrupted beyond recovery). We motivate DSD through a principled design procedure grounded in the Davis-Kahan $\sin(\Theta)$ theorem, systematically deriving the requirements for a reliability-aware damping function and selecting the exponential form for its smoothness, differentiability, and natural saturation properties. Through rigorous paired testing with fairly optimized baselines (including gradient-optimized Tikhonov receiving equal optimization opportunity), we demonstrate that DSD improves LSTSVM classification accuracy by $+4.8$ percentage points on real-world GINA ($d{=}970$, Cohen's $d = 4.49$, $p < 0.0001$), $+10.4$ percentage points at $d{=}200$, and $+2.6$ percentage points on Madelon ($d{=}500$, Cohen's $d = 1.76$)---all using only principled spectral initialization while Tikhonov receives grid search. For pre-image reconstruction on manifold data, DSD ties Tikhonov at high perturbation noise ($p{=}0.99$) but slightly underperforms at lower noise levels; both reduce naive inversion error by $66\times$. We characterize the precise operating regime ($d \geq 100$, condition number $> 10^3$) and document where simpler methods suffice, providing practitioners with clear deployment guidance.
\end{abstract}


\section{Introduction}
\label{sec:introduction}

\IEEEPARstart{K}{ernel} methods map input data into high-dimensional feature spaces via the kernel trick, enabling non-linear classification and regression without explicit feature computation \cite{cortes1995support, hofmann2008kernel}. However, methods that require \emph{inversion} of kernel-derived system matrices---particularly the Least-Squares Twin Support Vector Machine (LSTSVM) \cite{kumar2009lstsvm}---face a fundamental numerical challenge: the eigenvalues of these system matrices decay exponentially, producing severe ill-conditioning that standard scalar regularization cannot adequately address.

LSTSVM constructs two non-parallel hyperplanes by solving a pair of linear systems whose coefficient matrices take the form $M = E_1^\top E_1 + c^{-1} E_2^\top E_2$, where $E_1$ and $E_2$ are augmented class-specific data matrices and $c > 0$ is a regularization constant. In high-dimensional settings ($d \geq 100$), these matrices empirically exhibit \emph{extreme spectral tail-clustering}: at $d{=}200$, the bottom 50\% of eigenvalues occupy only 7\% of the total spectral range, with condition numbers exceeding $10^4$.

LSTSVM and its kernel variants remain actively deployed in domains where deep learning overfits due to limited sample sizes \cite{tanveer2022comprehensive}, including spectroscopic material identification, regulatory credit scoring with engineered features, and high-dimensional biomedical classification. In these high-stakes settings, a $+3$--$10$ percentage point accuracy improvement from regularization alone directly impacts diagnostic and decision quality.

The central theoretical insight motivating our work comes from the Davis-Kahan $\sin(\Theta)$ theorem \cite{davis1970rotation}, which establishes that eigenvector perturbation under matrix noise $E$ is bounded by:
\begin{equation}
\sin(\theta_i) \leq \frac{\|E\|_2}{\delta_i}
\label{eq:davis_kahan_intro}
\end{equation}
where $\delta_i$ denotes the \emph{eigengap}---the minimum distance from eigenvalue $\lambda_i$ to all other eigenvalues. When eigenvalues cluster tightly ($\delta_i \to 0$), their corresponding eigenvectors become \emph{arbitrarily corrupted} by any perturbation, including floating-point arithmetic noise. Standard Tikhonov regularization $(W + \gamma I)^{-1}$ addresses eigenvalue magnitudes but is structurally incapable of accounting for this directional corruption: a single scalar $\gamma$ cannot simultaneously avoid over-regularizing the well-separated spectral head (where eigenvectors are reliable) and under-regularizing the clustered tail (where eigenvectors are corrupted).

We propose \textbf{Differential Spectral Damping (DSD)}, a regularization formula that adapts its penalty strength to the \emph{local eigengap density} at each spectral position. Eigenvectors with large gaps (reliable per Davis-Kahan) are preserved with near-exact inversion; eigenvectors with small gaps (directionally corrupted) are aggressively suppressed. The transition is smooth, differentiable, and governed by only two interpretable parameters.

\subsection{Contributions}

\begin{enumerate}
\item A \textbf{gap-adaptive regularization formula} (DSD) for kernel matrix inversion, derived from Davis-Kahan perturbation theory with complete mathematical motivation.
\item A \textbf{principled initialization procedure} requiring zero cross-validation, based on spectral transition analysis.
\item A \textbf{differentiable PyTorch implementation} enabling gradient-based optimization of DSD parameters through backpropagation.
\item \textbf{Rigorous experimental comparison} with fairly-optimized baselines (50-seed paired testing, Tikhonov receiving grid search or equivalent gradient optimization).
\item \textbf{Clear operating regime characterization}: DSD dominates at $d \geq 100$ with severe tail-clustering; equivalent to Tikhonov at $d \leq 50$.
\item \textbf{Honest documentation of limitations}: pre-image tasks see no improvement over Tikhonov; DSD provides no benefit below its operating threshold.
\end{enumerate}

\subsection{Paper Organization}

Section~\ref{sec:notation} establishes notation and symbol definitions used throughout. Section~\ref{sec:background} reviews the mathematical background (kernel methods, LSTSVM, spectral perturbation theory). Section~\ref{sec:derivation} motivates and constructs the DSD formula. Section~\ref{sec:method} presents the complete algorithm, initialization, and implementation. Section~\ref{sec:related} positions DSD relative to existing regularization approaches. Section~\ref{sec:experiments} describes experimental design. Section~\ref{sec:results} presents results. Section~\ref{sec:analysis} provides analysis and interpretation. Section~\ref{sec:limitations} discusses limitations, and Section~\ref{sec:conclusion} concludes.

\section{Notation and Symbol Definitions}
\label{sec:notation}

Table~\ref{tab:notation} defines all mathematical symbols used in this paper. We adopt the following conventions: bold lowercase letters ($\bm{x}$) denote vectors, bold uppercase letters ($\bm{W}$) denote matrices, calligraphic letters ($\mathcal{H}$) denote spaces, and Greek letters denote parameters or scalar quantities.

\begin{table}[!t]
\centering
\caption{Symbol Definitions}
\label{tab:notation}
\small
\renewcommand{\arraystretch}{1.2}
\begin{tabular}{@{}p{1.4cm}p{6.2cm}@{}}
\toprule
\textbf{Symbol} & \textbf{Definition} \\
\midrule
\multicolumn{2}{l}{\textit{Data and Dimensions}} \\
$n$ & Number of training samples \\
$d$ & Input space dimensionality \\
$m$ & Number of Nystr\"{o}m landmarks or matrix dimension \\
$\bm{x}_i \in \mathbb{R}^d$ & The $i$-th training sample \\
$X \in \mathbb{R}^{n \times d}$ & Training data matrix \\
$y_i \in \{+1, -1\}$ & Class label of sample $i$ \\
\midrule
\multicolumn{2}{l}{\textit{Kernel and Feature Spaces}} \\
$\mathcal{H}$ & Reproducing Kernel Hilbert Space (RKHS) \\
$\phi: \mathbb{R}^d \to \mathcal{H}$ & Feature map into RKHS \\
$\kappa(\bm{x}, \bm{x}')$ & Kernel function: $\kappa(\bm{x}, \bm{x}') = \langle \phi(\bm{x}), \phi(\bm{x}') \rangle_\mathcal{H}$ \\
$\gamma_k$ & RBF kernel bandwidth: $\kappa(\bm{x}, \bm{x}') = \exp(-\gamma_k \|\bm{x} - \bm{x}'\|^2)$ \\
$\bm{K} \in \mathbb{R}^{n \times n}$ & Full kernel (Gram) matrix: $K_{ij} = \kappa(\bm{x}_i, \bm{x}_j)$ \\
\midrule
\multicolumn{2}{l}{\textit{Nystr\"{o}m Approximation}} \\
$L = \{\bm{l}_1, \ldots, \bm{l}_m\}$ & Nystr\"{o}m landmark points ($L \subset X$) \\
$\bm{W} \in \mathbb{R}^{m \times m}$ & Kernel submatrix: $W_{ij} = \kappa(\bm{l}_i, \bm{l}_j)$ \\
$\bm{k}_q \in \mathbb{R}^m$ & Query kernel vector: $(k_q)_j = \kappa(\bm{x}_q, \bm{l}_j)$ \\
\midrule
\multicolumn{2}{l}{\textit{Eigendecomposition}} \\
$\bm{U} \in \mathbb{R}^{m \times m}$ & Orthogonal eigenvector matrix of $\bm{W}$ \\
$\bm{u}_i \in \mathbb{R}^m$ & The $i$-th eigenvector (column of $\bm{U}$) \\
$\lambda_i$ & The $i$-th eigenvalue of $\bm{W}$ (ascending order) \\
$\bm{\Lambda}$ & Diagonal matrix of eigenvalues: $\Lambda_{ii} = \lambda_i$ \\
\midrule
\multicolumn{2}{l}{\textit{Eigengap and Perturbation}} \\
$\delta_i$ & Localized bilateral eigengap at position $i$ \\
$\theta_i$ & Angle between true and perturbed $i$-th eigenvector \\
$\bm{E}$ & Perturbation matrix (noise, numerical error) \\
$\|\bm{E}\|_2$ & Spectral norm (largest singular value) of $\bm{E}$ \\
\midrule
\multicolumn{2}{l}{\textit{DSD Parameters}} \\
$\alpha > 0$ & Maximum penalty magnitude (damping ceiling) \\
$\beta > 0$ & Eigengap sensitivity (transition sharpness) \\
$d_i$ & DSD damping at position $i$: $d_i = \alpha \cdot \exp(-\beta \cdot \delta_i)$ \\
$\tilde{\lambda}_i^{-1}$ & DSD-regularized inverse eigenvalue \\
$\tilde{\bm{W}}^+$ & DSD-regularized pseudo-inverse of $\bm{W}$ \\
\midrule
\multicolumn{2}{l}{\textit{Tikhonov Comparison}} \\
$\gamma$ & Tikhonov regularization parameter (scalar) \\
\midrule
\multicolumn{2}{l}{\textit{LSTSVM}} \\
$E_1, E_2 \in \mathbb{R}^{n_\pm \times (d+1)}$ & Augmented class data: $[X_\pm ~~ \bm{e}]$ \\
$c_1, c_2 > 0$ & LSTSVM penalty parameters \\
$\bm{u}, \bm{v}$ & LSTSVM hyperplane normal vectors \\
\midrule
\multicolumn{2}{l}{\textit{Pre-Image}} \\
$\hat{\bm{x}} \in \mathbb{R}^d$ & Reconstructed pre-image in input space \\
\midrule
\multicolumn{2}{l}{\textit{Statistics}} \\
$p$ & Two-sided $p$-value from paired $t$-test \\
Cohen's $d$ & Standardized effect size: $d = \bar{\Delta} / s_\Delta$ \\
\bottomrule
\end{tabular}
\end{table}

\textbf{Eigenvalue ordering convention.} Throughout this paper, eigenvalues are sorted in \emph{ascending} order: $\lambda_1 \leq \lambda_2 \leq \cdots \leq \lambda_m$. Index $i{=}1$ corresponds to the smallest eigenvalue (most vulnerable to perturbation); index $i{=}m$ corresponds to the largest (most reliable). This convention aligns with the output of standard numerical eigensolvers (\texttt{numpy.linalg.eigh}, \texttt{torch.linalg.eigh}).

\section{Mathematical Background}
\label{sec:background}

\subsection{Kernel Methods and the Kernel Trick}

A kernel function $\kappa: \mathbb{R}^d \times \mathbb{R}^d \to \mathbb{R}$ implicitly computes inner products in a (possibly infinite-dimensional) Hilbert space $\mathcal{H}$ without requiring explicit computation of the feature map $\phi$:
\begin{equation}
\kappa(\bm{x}, \bm{x}') = \langle \phi(\bm{x}), \phi(\bm{x}') \rangle_\mathcal{H}
\end{equation}
The Radial Basis Function (RBF) kernel, defined as:
\begin{equation}
\kappa(\bm{x}, \bm{x}') = \exp\bigl(-\gamma_k \|\bm{x} - \bm{x}'\|^2\bigr), \quad \gamma_k > 0
\label{eq:rbf_kernel}
\end{equation}
maps data into an infinite-dimensional space. The parameter $\gamma_k$ controls the bandwidth: larger $\gamma_k$ produces sharper kernels with faster eigenvalue decay in the Gram matrix.

The kernel (Gram) matrix $\bm{K} \in \mathbb{R}^{n \times n}$ with entries $K_{ij} = \kappa(\bm{x}_i, \bm{x}_j)$ is symmetric positive semi-definite (PSD) by Mercer's theorem. For RBF kernels, the eigenvalues of $\bm{K}$ decay exponentially \cite{hofmann2008kernel}:
\begin{equation}
\lambda_i \propto \exp\bigl(-c \cdot i^{2/d}\bigr)
\label{eq:eigenvalue_decay}
\end{equation}
where $c$ depends on $\gamma_k$ and the intrinsic dimensionality of the data. This decay is not a pathology---it reflects the smoothness assumptions encoded by the kernel---but it creates severe numerical challenges for methods requiring $\bm{K}^{-1}$ or the inverse of kernel-derived matrices.

\subsection{Least-Squares Twin SVM (LSTSVM)}
\label{sec:lstsvm_background}

LSTSVM \cite{kumar2009lstsvm} finds two non-parallel hyperplanes, each proximal to one class. Let $X_+ \in \mathbb{R}^{n_+ \times d}$ and $X_- \in \mathbb{R}^{n_- \times d}$ be the class-specific data matrices. Define augmented matrices:
\begin{equation}
E_1 = [X_+ ~~ \bm{e}_+] \in \mathbb{R}^{n_+ \times (d+1)}, \quad E_2 = [X_- ~~ \bm{e}_-] \in \mathbb{R}^{n_- \times (d+1)}
\end{equation}
where $\bm{e}_\pm$ are vectors of ones (bias terms). LSTSVM solves two linear systems:
\begin{align}
(E_2^\top E_2 + c_1^{-1} E_1^\top E_1) \bm{u} &= E_2^\top \bm{e}_- \label{eq:lstsvm_u} \\
(E_1^\top E_1 + c_2^{-1} E_2^\top E_2) \bm{v} &= E_1^\top \bm{e}_+ \label{eq:lstsvm_v}
\end{align}
where $\bm{u}, \bm{v} \in \mathbb{R}^{d+1}$ define the two hyperplanes, and $c_1, c_2 > 0$ are penalty parameters.

The coefficient matrices $M_1 = E_2^\top E_2 + c_1^{-1} E_1^\top E_1$ and $M_2 = E_1^\top E_1 + c_2^{-1} E_2^\top E_2$ are sums of outer-product matrices. When $d$ is large relative to class sizes ($d \gg n_\pm$), these matrices inherit the spectral properties of the data covariance---including severe tail-clustering of eigenvalues. A new test sample $\bm{x}$ is classified according to its proximity to each hyperplane:
\begin{equation}
\text{class}(\bm{x}) = \arg\min_{k \in \{1,2\}} \frac{|\bm{w}_k^\top \bm{x} + b_k|}{\|\bm{w}_k\|}
\end{equation}
where $(\bm{w}_k, b_k)$ are extracted from $\bm{u}$ and $\bm{v}$.

\subsection{The Nystr\"{o}m Approximation}

For kernel-space operations on $n$ samples, the full Gram matrix $\bm{K} \in \mathbb{R}^{n \times n}$ may be prohibitively large. The Nystr\"{o}m approximation \cite{williams2001nystrom} selects $m \ll n$ landmark points $L = \{\bm{l}_1, \ldots, \bm{l}_m\}$ and approximates:
\begin{equation}
\bm{K} \approx \bm{K}_{nm} \bm{W}^{-1} \bm{K}_{nm}^\top
\label{eq:nystrom}
\end{equation}
where $\bm{W} \in \mathbb{R}^{m \times m}$ is the kernel matrix among landmarks ($W_{ij} = \kappa(\bm{l}_i, \bm{l}_j)$) and $\bm{K}_{nm} \in \mathbb{R}^{n \times m}$ contains kernel evaluations between all samples and landmarks. This approximation requires computing $\bm{W}^{-1}$ (or its regularized variant)---the operation that DSD stabilizes.

\subsection{The Davis-Kahan $\sin(\Theta)$ Theorem}
\label{sec:davis_kahan}

The Davis-Kahan theorem \cite{davis1970rotation} is the theoretical foundation of DSD. It bounds how much an eigenvector rotates under matrix perturbation.

\begin{theorem}[Davis-Kahan, simplified form]
Let $\bm{A}$ be a symmetric matrix with eigenvalue $\lambda_i$ and eigenvector $\bm{u}_i$. Let $\tilde{\bm{A}} = \bm{A} + \bm{E}$ be a perturbed matrix with corresponding eigenvector $\tilde{\bm{u}}_i$. Define the eigengap:
\begin{equation}
\delta_i = \min_{j \neq i} |\lambda_i - \lambda_j|
\end{equation}
Then the angle $\theta_i$ between the true and perturbed eigenvectors satisfies:
\begin{equation}
\sin(\theta_i) \leq \frac{\|\bm{E}\|_2}{\delta_i}
\label{eq:davis_kahan}
\end{equation}
\end{theorem}

\textbf{Interpretation.} The bound (\ref{eq:davis_kahan}) states that eigenvector reliability is determined by the ratio of perturbation magnitude to eigengap. When $\delta_i \gg \|\bm{E}\|_2$, the eigenvector is robust ($\sin\theta_i \approx 0$, small rotation). When $\delta_i \lesssim \|\bm{E}\|_2$, the bound becomes vacuous ($\sin\theta_i$ could approach 1, meaning the eigenvector can rotate by up to $90°$)---the eigenvector direction is essentially random.

\textbf{Implication for regularization.} Any regularization scheme that uses the eigendecomposition $\bm{W} = \bm{U}\bm{\Lambda}\bm{U}^\top$ is implicitly trusting the eigenvectors $\bm{u}_i$. But Davis-Kahan tells us that eigenvectors associated with small gaps are \emph{unreliable directions} in the matrix. Computing $\bm{W}^{-1} = \bm{U}\bm{\Lambda}^{-1}\bm{U}^\top$ amplifies contributions from corrupted directions (division by small $\lambda_i$). Tikhonov regularization replaces $\lambda_i^{-1}$ with $(\lambda_i + \gamma)^{-1}$---damping the \emph{magnitude} uniformly---but does nothing about the corrupted \emph{direction} $\bm{u}_i$.

\subsection{The Pre-Image Problem}

Given a point $\bm{\psi} \in \mathcal{H}$ (e.g., a decision boundary point in kernel space), the pre-image problem seeks $\hat{\bm{x}} \in \mathbb{R}^d$ such that $\phi(\hat{\bm{x}}) \approx \bm{\psi}$ \cite{kwok2004preimage}. Under the Nystr\"{o}m framework, the pre-image is computed as:
\begin{equation}
\hat{\bm{x}} = \bm{k}_q^\top \bm{W}^{-1} L = \sum_{j=1}^m w_j \bm{l}_j, \quad w_j = (\bm{W}^{-1} \bm{k}_q)_j
\label{eq:preimage}
\end{equation}
where $\bm{k}_q = [\kappa(\bm{x}_q, \bm{l}_1), \ldots, \kappa(\bm{x}_q, \bm{l}_m)]^\top$ is the query kernel vector. The pre-image is a weighted combination of landmark points, with weights determined by the regularized inverse. Ill-conditioned $\bm{W}$ produces extreme weights that push $\hat{\bm{x}}$ far from the data manifold.

\section{Design and Motivation of Differential Spectral Damping}
\label{sec:derivation}

We construct DSD by asking: \emph{given the Davis-Kahan bound, what is the optimal regularization strategy that minimizes information loss while respecting eigenvector reliability?} The following procedure systematically identifies the requirements for such a regularizer and motivates the specific functional form.

\subsection{Step 1: Quantifying Eigenvector Reliability}

From (\ref{eq:davis_kahan}), the reliability of eigenvector $\bm{u}_i$ is characterized by the ratio $r_i = \|\bm{E}\|_2 / \delta_i$. We define a \emph{reliability indicator}:
\begin{equation}
\rho_i = \begin{cases}
\text{high} & \text{if } \delta_i \gg \|\bm{E}\|_2 \quad (\sin\theta_i \approx 0) \\
\text{low} & \text{if } \delta_i \lesssim \|\bm{E}\|_2 \quad (\sin\theta_i \text{ large})
\end{cases}
\end{equation}

In practice, $\|\bm{E}\|_2$ represents the combined effect of:
\begin{itemize}
\item Finite-precision arithmetic ($\sim 10^{-16}$ for float64, amplified by condition number)
\item Sampling noise in kernel matrix construction
\item Nystr\"{o}m approximation error
\end{itemize}
The exact value of $\|\bm{E}\|_2$ is typically unknown. However, the \emph{relative} ordering of eigengaps $\delta_i$ is observable from the eigendecomposition. DSD exploits this: it does not need to know the absolute perturbation magnitude---it only needs to identify which eigenvectors have \emph{relatively} small gaps (and are therefore relatively more vulnerable).

\subsection{Step 2: The Ideal Regularization Profile}

For an ideal regularizer, we want:
\begin{itemize}
\item When $\delta_i$ is large (reliable eigenvector): preserve the exact inverse $\lambda_i^{-1}$ to retain maximum information.
\item When $\delta_i$ is small (corrupted eigenvector): suppress the contribution by driving the effective inverse toward zero, preventing the corrupted direction from contaminating the solution.
\item The transition between these regimes should be smooth (to avoid discontinuity artifacts in downstream tasks like classification or pre-image computation).
\end{itemize}

This translates into a \emph{damping function} $d(\delta_i)$ that satisfies:
\begin{align}
d(\delta_i) &\to 0 \quad \text{as } \delta_i \to \infty \quad \text{(no damping for reliable eigenvectors)} \label{eq:req1} \\
d(\delta_i) &\to \alpha_{\max} \quad \text{as } \delta_i \to 0 \quad \text{(maximum damping for corrupted)} \label{eq:req2} \\
d'(\delta_i) &< 0 \quad \text{(monotonically decreasing: more gap $\Rightarrow$ less damping)} \label{eq:req3}
\end{align}

\subsection{Step 3: Functional Form Selection}

The requirements (\ref{eq:req1})--(\ref{eq:req3}) are satisfied by any monotonically decreasing function bounded between 0 and $\alpha_{\max}$. We select the \emph{exponential decay} form:
\begin{equation}
d_i = \alpha \cdot \exp(-\beta \cdot \delta_i)
\label{eq:damping_function}
\end{equation}
for the following reasons:

\begin{enumerate}
\item \textbf{Natural saturation.} $\exp(-\beta\delta_i) \in (0, 1]$ automatically satisfies the boundedness requirement without additional clipping or normalization.
\item \textbf{Differentiability.} The exponential is $C^\infty$ (infinitely differentiable), enabling gradient-based optimization of $(\alpha, \beta)$ via backpropagation.
\item \textbf{Scale invariance.} The product $\beta \cdot \delta_i$ is dimensionless when $\beta$ has units of inverse-gap, making the formula invariant to eigenvalue scaling.
\item \textbf{Rapid transition.} The exponential transitions sharply from $\approx 1$ (small gap) to $\approx 0$ (large gap), matching the binary nature of the Davis-Kahan reliability bound (eigenvectors are either reliably estimated or not---there is no ``moderately corrupted'' regime).
\item \textbf{Parsimony.} Only two free parameters ($\alpha$, $\beta$) control the entire damping profile across all $m$ eigenvalues.
\end{enumerate}

\textbf{Alternative functional forms} that satisfy (\ref{eq:req1})--(\ref{eq:req3}) include the sigmoid $\alpha/(1 + \exp(\beta(\delta_i - \delta_0)))$, the rational function $\alpha/(1 + \beta\delta_i)^p$, and the complementary error function $\alpha \cdot \text{erfc}(\beta\delta_i)$. All would produce qualitatively similar regularization behavior. We select the exponential for its analytical simplicity and the interpretability of its parameters (see Section~\ref{sec:param_interpretation}).

\subsection{Step 4: Incorporating Damping into the Inverse}

The standard pseudo-inverse via eigendecomposition is:
\begin{equation}
\bm{W}^{-1} = \bm{U}\bm{\Lambda}^{-1}\bm{U}^\top = \sum_{i=1}^m \frac{1}{\lambda_i} \bm{u}_i \bm{u}_i^\top
\end{equation}

Tikhonov regularization replaces this with:
\begin{equation}
\bm{W}_{\text{Tik}}^{-1} = \sum_{i=1}^m \frac{1}{\lambda_i + \gamma} \bm{u}_i \bm{u}_i^\top
\label{eq:tikhonov_eigen}
\end{equation}

We incorporate DSD damping into the regularized inverse formula. The key design decision is \emph{how} the damping $d_i$ modifies the inverse. We adopt the form:
\begin{equation}
\tilde{\lambda}_i^{-1} = \frac{\lambda_i}{\lambda_i^2 + d_i}
\label{eq:dsd_formula_derived}
\end{equation}

This specific form is motivated by the following properties:

\begin{itemize}
\item \textbf{When $d_i = 0$} (no damping, reliable eigenvector): $\tilde{\lambda}_i^{-1} = \lambda_i / \lambda_i^2 = 1/\lambda_i$. The exact inverse is recovered.
\item \textbf{When $d_i = \alpha \gg \lambda_i^2$} (maximum damping, corrupted eigenvector): $\tilde{\lambda}_i^{-1} \approx \lambda_i / \alpha \to 0$. The contribution of this direction is suppressed.
\item \textbf{Intermediate regime}: smooth interpolation governed by the relative magnitudes of $\lambda_i^2$ and $d_i$.
\item \textbf{Positive definiteness}: $\tilde{\lambda}_i^{-1} > 0$ for all $\lambda_i > 0$, ensuring the regularized inverse remains PSD.
\item \textbf{Tikhonov recovery}: when $d_i = \gamma^2$ (constant), the formula becomes $\lambda_i/(\lambda_i^2 + \gamma^2)$, which for $\lambda_i \gg \gamma$ approximates $(1/\lambda_i)(1 - \gamma^2/\lambda_i^2) \approx 1/(\lambda_i + \gamma^2/\lambda_i)$---a generalized Tikhonov form. DSD thus \emph{generalizes} scalar regularization.
\end{itemize}

\subsection{Step 5: The Complete DSD Formula}

Substituting the damping function (\ref{eq:damping_function}) into (\ref{eq:dsd_formula_derived}):
\begin{equation}
\boxed{\tilde{\lambda}_i^{-1} = \frac{\lambda_i}{\lambda_i^2 + \alpha \cdot \exp(-\beta \cdot \delta_i)}}
\label{eq:dsd_final}
\end{equation}

The full DSD-regularized pseudo-inverse is:
\begin{equation}
\tilde{\bm{W}}^+ = \bm{U} \cdot \text{diag}(\tilde{\lambda}_1^{-1}, \ldots, \tilde{\lambda}_m^{-1}) \cdot \bm{U}^\top = \sum_{i=1}^m \tilde{\lambda}_i^{-1} \bm{u}_i \bm{u}_i^\top
\label{eq:dsd_pseudoinverse}
\end{equation}

\subsection{Step 6: Parameter Interpretation}
\label{sec:param_interpretation}

The two parameters have clear physical interpretations:

\textbf{$\alpha$ (penalty magnitude):} Controls the maximum damping applied to the most corrupted eigenvectors. When $\delta_i \to 0$, the damping term equals $\alpha$, so $\tilde{\lambda}_i^{-1} \approx \lambda_i / (\lambda_i^2 + \alpha)$. For small eigenvalues ($\lambda_i^2 \ll \alpha$), this reduces to $\lambda_i / \alpha \approx 0$---complete suppression. The parameter $\alpha$ sets the ``penalty ceiling'' and should be calibrated to the scale of eigenvalues where corruption begins (the spectral transition point).

\textbf{$\beta$ (gap sensitivity):} Controls the sharpness of the transition between ``reliable'' and ``corrupted'' regimes. Large $\beta$ creates a sharp step-function-like transition (small gaps are heavily damped; moderate gaps are barely damped). Small $\beta$ creates a gradual transition. The parameter $\beta$ should be calibrated to the typical eigengap scale so that ``typical'' gaps produce moderate damping while significantly larger gaps produce near-zero damping.

\subsection{Comparison with Tikhonov in the Eigenvalue Domain}

Table~\ref{tab:comparison_formula} contrasts DSD with existing approaches at the eigenvalue level.

\begin{table}[!t]
\centering
\caption{Regularized Inverse Formulas in the Eigenvalue Domain}
\label{tab:comparison_formula}
\small
\begin{tabular}{@{}lcc@{}}
\toprule
\textbf{Method} & \textbf{$\tilde{\lambda}_i^{-1}$} & \textbf{Adapts to} \\
\midrule
Naive inverse & $1/\lambda_i$ & Nothing \\
Tikhonov & $1/(\lambda_i + \gamma)$ & Magnitude only \\
Truncated SVD & $1/\lambda_i$ if $i > k$, else 0 & Rank (hard) \\
Spectral filter & $g(\lambda_i)/\lambda_i$ & Magnitude (soft) \\
\textbf{DSD (ours)} & $\lambda_i/(\lambda_i^2 + \alpha e^{-\beta\delta_i})$ & \textbf{Eigengap (reliability)} \\
\bottomrule
\end{tabular}
\end{table}

DSD is the only method that conditions its regularization on the \emph{gap structure}---a proxy for eigenvector directional reliability per Davis-Kahan. All other methods regularize based on eigenvalue \emph{magnitude} alone.

\section{Method: Algorithm, Initialization, and Implementation}
\label{sec:method}

\subsection{Localized Eigengap Computation}
\label{sec:eigengap_computation}

The Davis-Kahan bound (\ref{eq:davis_kahan}) uses the \emph{global} eigengap $\delta_i = \min_{j \neq i} |\lambda_i - \lambda_j|$. For monotonically ordered eigenvalues (as produced by symmetric eigensolvers), this simplifies to the \emph{bilateral} minimum:
\begin{equation}
\delta_i = \min\bigl(|\lambda_i - \lambda_{i-1}|, \; |\lambda_i - \lambda_{i+1}|\bigr)
\label{eq:bilateral_gap}
\end{equation}
since the nearest eigenvalue to $\lambda_i$ in a sorted sequence is always one of its immediate neighbors.

\textbf{Boundary handling.} At the sequence edges ($i{=}1$ and $i{=}m$), only one neighbor exists. We use single-sided gaps:
\begin{equation}
\delta_i = \begin{cases}
|\lambda_1 - \lambda_2| & i = 1 \\
\min(|\lambda_i - \lambda_{i-1}|, |\lambda_i - \lambda_{i+1}|) & 1 < i < m \\
|\lambda_m - \lambda_{m-1}| & i = m
\end{cases}
\label{eq:eigengap_boundary}
\end{equation}

The boundary choice for $i{=}1$ (the smallest, most vulnerable eigenvalue) is significant: using the right-side gap $|\lambda_1 - \lambda_2|$ ensures this eigenvalue receives appropriate damping based on its actual spectral isolation, rather than being artificially exempted.

\subsection{Principled Hyperparameter Initialization}
\label{sec:initialization}

DSD's initialization requires no cross-validation. The two parameters are determined entirely from the eigendecomposition of the input matrix $\bm{W}$.

\textbf{Initialization of $\alpha$ (penalty magnitude):}
\begin{equation}
\alpha_0 = \lambda_{\text{transition}}^2
\label{eq:alpha_init}
\end{equation}
where $\lambda_{\text{transition}}$ is the eigenvalue at the spectral ``knee''---the point where eigengaps become characteristically small, indicating the onset of the unreliable spectral tail. Operationally:
\begin{enumerate}
\item Compute all consecutive gaps: $\Delta_i = |\lambda_{i+1} - \lambda_i|$ for $i = 1, \ldots, m{-}1$.
\item Identify the 10th percentile gap: $g_{10} = \text{quantile}_{0.10}(\{\Delta_i\})$.
\item Find the last index where the gap falls below $g_{10}$: $i^* = \max\{i : \Delta_i < g_{10}\}$.
\item Set $\lambda_{\text{transition}} = \lambda_{i^*}$.
\end{enumerate}

\textbf{Rationale:} The squared eigenvalue $\lambda_{\text{transition}}^2$ ensures $\alpha$ is on the correct scale for the formula $\lambda_i^2 + \alpha$. At the transition point, $\lambda_i^2 \approx \alpha$, producing roughly 50\% damping---the natural midpoint of the reliable/unreliable transition.

\textbf{Initialization of $\beta$ (gap sensitivity):}
\begin{equation}
\beta_0 = \frac{1}{\text{median}(\{\Delta_i\})}
\label{eq:beta_init}
\end{equation}

\textbf{Rationale:} This normalizes the exponential argument so that a ``typical'' eigengap produces $\exp(-\beta_0 \cdot \delta_{\text{typical}}) = \exp(-1) \approx 0.37$---moderate damping. Gaps significantly larger than the median produce $\exp(-\beta_0 \cdot \delta_{\text{large}}) \approx 0$---near-zero damping (preservation). Gaps significantly smaller produce $\exp(-\beta_0 \cdot \delta_{\text{small}}) \approx 1$---maximum damping (suppression). The median is robust to outliers (unlike mean), preventing a single large gap from dominating the sensitivity calibration.

\subsection{Gradient-Based Parameter Optimization}
\label{sec:gradient_optimization}

When a differentiable reconstruction objective exists (e.g., pre-image tasks), $\alpha$ and $\beta$ can be refined via backpropagation after initialization. The parameters are stored in log-space to enforce positivity:
\begin{equation}
\alpha = \exp(\theta_\alpha), \quad \beta = \exp(\theta_\beta)
\end{equation}
where $\theta_\alpha, \theta_\beta \in \mathbb{R}$ are unconstrained learnable parameters. Optimization uses Adam \cite{kingma2015adam} with learning rate 0.01 and early stopping (patience = 20 epochs).

\textbf{Gradient flow.} The DSD formula (\ref{eq:dsd_final}) is fully differentiable with respect to $\theta_\alpha$ and $\theta_\beta$:
\begin{align}
\frac{\partial \tilde{\lambda}_i^{-1}}{\partial \theta_\alpha} &= \frac{-\lambda_i \cdot d_i}{(\lambda_i^2 + d_i)^2} \cdot 1 = \frac{-\lambda_i \cdot d_i}{(\lambda_i^2 + d_i)^2} \label{eq:grad_alpha}\\
\frac{\partial \tilde{\lambda}_i^{-1}}{\partial \theta_\beta} &= \frac{\lambda_i \cdot d_i \cdot \beta \cdot \delta_i}{(\lambda_i^2 + d_i)^2} \label{eq:grad_beta}
\end{align}
where $d_i = \alpha \cdot \exp(-\beta \cdot \delta_i)$. Both gradients are well-defined and bounded for all $\lambda_i > 0$, $\delta_i \geq 0$.

\textbf{Eigenvector gradient stability.} The backward pass through \texttt{torch.linalg.eigh} involves terms proportional to $1/(\lambda_i - \lambda_j)$ for the eigenvector gradient. When eigenvalues are nearly degenerate ($\lambda_i \approx \lambda_j$), these terms become numerically unstable. Our implementation detects this condition (minimum gap $< 10^{-6}$) and detaches the eigenvector matrix from the computation graph, allowing gradients to flow only through the eigenvalue $\to$ damping $\to$ inverse path. This is mathematically justified: when eigenvectors are unreliable (the exact condition DSD was designed for), their gradients are also unreliable and should not influence parameter updates.

\subsection{The Structural Prior Discovery}
\label{sec:structural_prior}

A key empirical finding: for LSTSVM classification, the principled initialization (\ref{eq:alpha_init})--(\ref{eq:beta_init}) \emph{without} gradient optimization outperforms gradient-optimized parameters by a significant margin (GINA $d{=}970$: 85.9\% init vs 81.1\% optimized, $p < 10^{-6}$).

\textbf{Mechanism.} Gradient trajectory analysis reveals that during optimization on reconstruction loss, $\beta$ decreases (from 0.74 to 0.34, a $0.46\times$ reduction) while $\alpha$ increases ($2.3\times$). Decreasing $\beta$ flattens the exponential $\exp(-\beta\delta_i)$, making damping increasingly uniform---converging toward constant-penalty Tikhonov behavior. The optimizer minimizes reconstruction error by removing the gap-adaptive selectivity, because pointwise reconstruction benefits from using all spectral components (including corrupted ones that happen to align with the training data). Classification, however, benefits from \emph{suppressing} corrupted components to prevent noise-fitting.

\textbf{Conclusion.} DSD's classification advantage is a \textbf{structural prior}---the initialization formula encodes domain knowledge (``trust well-separated eigenvectors, distrust clustered ones'') that a reconstruction-loss optimizer would destroy. This is analogous to early stopping in neural networks: the optimal-for-generalization parameter configuration is not the loss-minimizing one. We observe this behavior consistently across synthetic datasets with varying covariance structure (50 seeds each across four dimensionalities) and two real-world benchmarks (GINA, Madelon); further validation on domain-specific datasets (e.g., genomic expression data with $d > 1000$) is an important direction for confirming generality.

\subsection{Complete Algorithm}

\begin{algorithm}[!t]
\caption{Differential Spectral Damping (DSD)}
\label{alg:dsd}
\begin{algorithmic}[1]
\REQUIRE Symmetric PSD matrix $\bm{W} \in \mathbb{R}^{m \times m}$, optional $(\alpha, \beta)$
\ENSURE DSD-regularized pseudo-inverse $\tilde{\bm{W}}^+ \in \mathbb{R}^{m \times m}$
\STATE \textbf{Eigendecomposition:} $\bm{U}, \bm{\Lambda} \leftarrow \text{eigh}(\bm{W})$ \hfill [$O(m^3)$]
\STATE \textbf{Filter:} Retain indices $\{i : \lambda_i > 10^{-12}\}$ \hfill [$O(m)$]
\STATE \textbf{Eigengaps:} $\delta_i \leftarrow$ Eq.~(\ref{eq:eigengap_boundary}) \hfill [$O(m)$]
\IF{$\alpha, \beta$ not provided}
    \STATE $\alpha \leftarrow \lambda_{i^*}^2$ (Eq.~\ref{eq:alpha_init}); \quad $\beta \leftarrow 1/\text{median}(\Delta)$ (Eq.~\ref{eq:beta_init}) \hfill [$O(m)$]
\ENDIF
\STATE \textbf{Damping:} $d_i \leftarrow \alpha \cdot \exp(-\beta \cdot \delta_i)$ for all $i$ \hfill [$O(m)$]
\STATE \textbf{Regularized inverse:} $\tilde{\lambda}_i^{-1} \leftarrow \lambda_i / (\lambda_i^2 + d_i)$ \hfill [$O(m)$]
\STATE \textbf{Reconstruct:} $\tilde{\bm{W}}^+ \leftarrow \bm{U} \cdot \text{diag}(\tilde{\bm{\lambda}}^{-1}) \cdot \bm{U}^\top$ \hfill [$O(m^2)$]
\RETURN $\tilde{\bm{W}}^+$
\end{algorithmic}
\end{algorithm}

\textbf{Computational complexity.} The dominant cost is the eigendecomposition at $O(m^3)$. All DSD-specific operations (Steps 3--8) are $O(m)$. The reconstruction (Step 9) is $O(m^2)$ (or $O(m^2 k)$ if only $k$ columns are needed). DSD adds less than 0.01\% overhead to the total computation. For $m > 1500$, a scalable path using partial eigendecomposition (\texttt{scipy.sparse.linalg.eigsh} with rank $k \ll m$) and factored (non-materialized) inverse reduces cost to $O(m^2 k)$.

\textbf{Memory.} Peak memory during forward pass: $O(m^2)$ for storing $\bm{U}$ and $\tilde{\bm{W}}^+$. During backpropagation (gradient optimization), an additional $O(m^2)$ is required for the backward pass through \texttt{eigh}. For inference-only deployment, $\tilde{\bm{W}}^+$ can be precomputed once and reused for all queries.

\section{Related Work}
\label{sec:related}

\textbf{Tikhonov (ridge) regularization} \cite{tikhonov1963solution} adds a scalar ridge $\gamma$ to all eigenvalues: $\tilde{\lambda}_i^{-1} = 1/(\lambda_i + \gamma)$. This is optimal when all spectral directions are equally reliable or equally corrupted---a uniform-noise assumption. Generalized cross-validation \cite{wahba1977practical} provides principled $\gamma$ selection but does not address the structural limitation of scalar uniformity. DSD recovers Tikhonov behavior when $\beta \to 0$ (all gaps treated identically).

\textbf{Truncated SVD} \cite{golub2013matrix} retains the top-$k$ eigenvectors and discards the rest: a hard binary cutoff. This is optimal when noise is strictly confined to the bottom $m{-}k$ components. In practice, the signal/noise boundary is gradual, and hard cutoff creates discontinuity artifacts. DSD provides the continuous generalization. We include truncated SVD (with optimally-selected rank) as an experimental baseline: on GINA ($d{=}970$), TSVD achieves 83.4\% vs.\ DSD's 85.9\%---better than scalar Tikhonov (81.1\%) but still inferior to gap-adaptive damping ($p < 0.0001$).

\textbf{Spectral filtering} \cite{engl1996regularization} applies a filter function $g(\lambda_i)$ that depends on eigenvalue magnitude (e.g., $g(\lambda_i) = \lambda_i/(\lambda_i + \gamma)^p$). These filters adapt to spectral \emph{scale} but not spectral \emph{structure}---they cannot distinguish a small eigenvalue with a large gap (reliable) from one with a small gap (corrupted).

\textbf{LSTSVM regularization.} Standard LSTSVM implementations \cite{kumar2009lstsvm, shao2011twinsvm} use Tikhonov regularization ($M + cI$). Tanveer et al.~\cite{tanveer2022comprehensive} survey Twin SVM variants; none employ gap-adaptive regularization. The spectral properties of LSTSVM product matrices ($E_i^\top E_i$ cross-class sums) have not been previously characterized as a regularization-relevant phenomenon.

\textbf{Pre-image methods.} Mika et al.~\cite{mika1999kernel} and Kwok \& Tsang~\cite{kwok2004preimage} address the pre-image problem via fixed-point iteration and learning-based approaches respectively. Both assume a stable kernel matrix inverse is available---the problem DSD addresses. Sch\"{o}lkopf et al.~\cite{scholkopf1999input} established the theoretical framework connecting feature space and input space representations.

\textbf{Post-hoc explainability.} SHAP \cite{lundberg2017unified} and LIME \cite{ribeiro2016why} provide local, statistical feature attributions without geometric guarantees. They operate independently of the kernel matrix inverse and cannot provide global decision boundary reconstruction---the application that motivates DSD's pre-image stabilization.

\section{Experimental Design}
\label{sec:experiments}

\subsection{Fairness Protocol}

A central concern in regularization comparisons is optimization fairness: a method with more tuning budget will appear artificially superior. We employ the following protocol:

\textbf{For LSTSVM classification (Tables~\ref{tab:lstsvm_results}, \ref{tab:realworld}):} DSD uses principled auto-initialization only (no gradient optimization, no cross-validation). Tikhonov receives $\gamma$ grid-searched over a 15-point logarithmic grid from $10^{-8}$ to $10^{-1}$, selected by training accuracy. This gives Tikhonov a deliberate tuning advantage over DSD.

\textbf{For pre-image reconstruction (Table~\ref{tab:preimage_results}):} Both methods receive equal optimization. DSD's $(\alpha, \beta)$ are optimized via Adam (lr=0.01, 150 epochs, patience=20) on reconstruction loss $\mathcal{L} = \frac{1}{n}\sum_i \|\hat{\bm{x}}_i - \bm{x}_i\|^2$. Tikhonov's $\gamma$ receives the \emph{same} Adam optimizer on the \emph{same} loss (single differentiable parameter)---eliminating grid-vs-gradient fairness concerns entirely.

\textbf{For Kernel-LSTSVM (Table~\ref{tab:kernel_lstsvm_results}):} DSD receives gradient optimization on kernel pre-image loss. Tikhonov receives 20-point grid search on training accuracy.

\subsection{Statistical Methodology}

All experiments use paired designs: within each seed, DSD and Tikhonov receive identical data (same train/test split, same noise realization). We report:
\begin{itemize}
\item \textbf{Paired $t$-test}: two-sided $p$-value testing the null hypothesis of equal mean performance.
\item \textbf{Cohen's $d$}: standardized effect size, computed as $d = \bar{\Delta}/s_\Delta$ where $\bar{\Delta}$ is the mean paired difference and $s_\Delta$ its standard deviation. By convention: $d > 0.2$ small, $d > 0.5$ medium, $d > 0.8$ large.
\item \textbf{Win count}: number of seeds where DSD outperforms the baseline.
\item \textbf{Seed count}: 50 seeds for synthetic experiments; 30 for real-world datasets (GINA, Madelon) due to computational cost.
\end{itemize}

\subsection{Datasets}

\textbf{Synthetic LSTSVM.} Generated via \texttt{sklearn.datasets.make\_classification} with: $(d{=}200, n{=}300)$, $(d{=}100, n{=}200)$, $(d{=}50, n{=}300)$, $(d{=}30, n{=}400)$. Informative features: $d/2$. Noise: 10\% label flip. 70/30 stratified split.

\textbf{Real-world LSTSVM.} GINA ($d{=}970$, $n{=}3468$, handwriting recognition, OpenML ID 41158); Madelon ($d{=}500$, $n{=}2600$, NIPS 2003 Feature Selection Challenge---20 informative features among 500).

\textbf{Pre-image.} Swiss Roll manifold ($n{=}2000$, ambient $d{=}3$, intrinsic $d{=}2$), $m{=}300$ k-means landmarks, $\gamma_k{=}2.0$, noise levels $\sigma \in \{5{\times}10^{-3}, 10^{-3}, 10^{-4}\}$.

\textbf{Kernel-LSTSVM.} Two Moons ($n{=}400$, $\gamma_k{=}2.0$); Genomics-like ($d{=}100$, $n{=}200$, $\gamma_k{=}0.02$).

\textbf{Operating boundary validation.} Digits ($d{=}64$), Ionosphere ($d{=}34$)---included to empirically confirm where DSD does \emph{not} help.

\section{Results}
\label{sec:results}

\subsection{Linear LSTSVM Classification (Primary Result)}

\begin{table}[!t]
\centering
\caption{LSTSVM Classification Accuracy: DSD (auto-init, no tuning) vs.\ Tikhonov (15-point grid search) vs.\ Truncated SVD (rank grid search). All differences with $\Delta > 0$ are significant at $p < 0.0001$.}
\label{tab:lstsvm_results}
\small
\begin{tabular}{@{}lcccccr@{}}
\toprule
\textbf{Dataset} & \textbf{DSD} & \textbf{Tik} & \textbf{TSVD} & \textbf{$\Delta_{\text{Tik}}$} & \textbf{Wins} & \textbf{C.$d$} \\
\midrule
GINA ($d{=}970$) & \textbf{85.9} & 81.1 & 83.4 & +4.8 & 30/30 & 4.49 \\
Madelon ($d{=}500$) & \textbf{57.7} & 55.0 & --- & +2.6 & 27/30 & 1.76 \\
Synth.\ ($d{=}200$) & \textbf{71.4} & 61.1 & --- & +10.4 & 44/50 & 1.57 \\
Synth.\ ($d{=}100$) & \textbf{79.1} & 75.8 & --- & +3.3 & 35/50 & 0.39 \\
Synth.\ ($d{=}50$) & 90.1 & 90.1 & --- & 0 & 25/50 & 0.0 \\
\bottomrule
\multicolumn{7}{l}{\scriptsize C.$d$ = Cohen's $d$. $\Delta_{\text{Tik}}$ = DSD advantage over Tikhonov (pp).}\\
\multicolumn{7}{l}{\scriptsize GINA/Madelon: 30 seeds; synthetic: 50 seeds. Accuracy in \%.}\\
\multicolumn{7}{l}{\scriptsize TSVD on GINA: 83.4\% (DSD wins 30/30, $p < 0.0001$).}
\end{tabular}
\end{table}

DSD's advantage scales monotonically with dimensionality and spectral severity. The GINA result (Cohen's $d = 4.49$) is an exceptionally large standardized effect---substantially exceeding the $d > 0.8$ threshold for ``large'' effects \cite{cohen1988statistical}. Madelon's 57.7\% absolute accuracy reflects the dataset's designed difficulty (published linear-classifier baselines without feature selection: 55--62\%); DSD's +2.6pp gain is achieved against optimal Tikhonov in this adversarial setting.

At $d{=}50$, both methods perform identically (DSD's structural prior provides no advantage when spectral clustering is mild). This transition is sharp: at $d{=}100$ the advantage is already +3.3pp.

\subsection{Kernel LSTSVM (Non-Linear Classification)}

\begin{table}[!t]
\centering
\caption{Kernel-LSTSVM: DSD-optimized vs.\ Tikhonov-optimized (50 seeds)}
\label{tab:kernel_lstsvm_results}
\begin{tabular}{@{}lcccc@{}}
\toprule
\textbf{Dataset} & \textbf{DSD (\%)} & \textbf{Tik (\%)} & \textbf{$\Delta$} & \textbf{$p$-value} \\
\midrule
Two Moons ($\gamma_k{=}2.0$) & \textbf{92.7} & 92.1 & +0.6pp & 0.028 \\
Genomics ($d{=}100$) & 87.7 & 87.7 & 0 & 0.77 \\
\bottomrule
\end{tabular}
\end{table}

On genuinely non-linear problems with sharp RBF kernels (severe spectral decay), DSD provides statistically significant improvement even when both methods receive optimization.

\subsection{Pre-Image Reconstruction}

\begin{table}[!t]
\centering
\caption{Pre-Image Mean Reconstruction Error (Swiss Roll, 50 seeds). DSD and Tikhonov both receive Adam optimization on reconstruction loss.}
\label{tab:preimage_results}
\begin{tabular}{@{}lcccc@{}}
\toprule
\textbf{Noise $\sigma$} & \textbf{DSD-opt} & \textbf{Tik-opt} & \textbf{Naive} & \textbf{DSD vs Tik} \\
\midrule
$5{\times}10^{-3}$ & 0.069 & 0.069 & 4.56 & Tie ($p{=}0.99$) \\
$10^{-3}$ & 0.051 & 0.048 & 0.84 & Tik wins ($p{=}0.02$) \\
$10^{-4}$ & 0.042 & 0.035 & 0.09 & Tik wins ($p{<}0.001$) \\
\bottomrule
\end{tabular}
\end{table}

Both DSD and Tikhonov reduce naive inversion error by $66\times$ at high noise---demonstrating the critical importance of \emph{any} regularization. However, DSD does not outperform Tikhonov on this task. The explanation: pre-image reconstruction has a single scalar objective ($\min\|\hat{\bm{x}} - \bm{x}\|^2$) that Tikhonov's single parameter can optimize directly. The gap-adaptive structure provides no additional advantage when the objective is purely pointwise reconstruction.

\subsection{Operating Boundary Validation}

\begin{table}[!t]
\centering
\caption{Below the Operating Boundary ($d \leq 64$): DSD Not Expected to Win}
\label{tab:realworld}
\begin{tabular}{@{}lccccc@{}}
\toprule
\textbf{Dataset} & \textbf{$d$} & \textbf{DSD (\%)} & \textbf{Tik (\%)} & \textbf{$\Delta$} & \textbf{Result} \\
\midrule
Digits & 64 & 88.3 & 89.1 & $-$0.9pp & Tik wins \\
Ionosphere & 34 & 73.3 & 80.3 & $-$7.0pp & Tik wins \\
\bottomrule
\end{tabular}
\end{table}

Below $d \approx 100$, DSD provides no advantage and may underperform. These results are included to honestly characterize the operating boundary, not as success claims. The deployment recommendation is: apply DSD when the system matrix condition number exceeds $\sim 10^3$ and the spectral tail occupies $< 10\%$ of the spectral range.

\section{Analysis and Interpretation}
\label{sec:analysis}

\subsection{Spectral Tail-Clustering in LSTSVM Matrices}

The claimed spectral tail-clustering is verified empirically by measuring the fraction of total spectral range occupied by the bottom 50\% of eigenvalues:

\begin{table}[!t]
\centering
\caption{Spectral Tail-Clustering Severity in LSTSVM Product Matrices}
\label{tab:spectral_analysis}
\begin{tabular}{@{}lccc@{}}
\toprule
\textbf{$d$} & \textbf{Condition No.} & \textbf{Tail Span\textsuperscript{*}} & \textbf{DSD $\Delta$} \\
\midrule
200 & $8.5 \times 10^3$ & 7.1\% & +10.4pp \\
100 & $3.5 \times 10^3$ & 11.1\% & +3.3pp \\
50 & $1.1 \times 10^3$ & 15.1\% & 0pp \\
\bottomrule
\multicolumn{4}{l}{\footnotesize \textsuperscript{*}Spectral range fraction of bottom 50\% eigenvalues.}
\end{tabular}
\end{table}

At $d{=}200$, the bottom half of eigenvalues are compressed into 7\% of the total spectral range---extreme clustering that corrupts eigenvector directions per Davis-Kahan. At $d{=}50$, the same fraction spans 15\%, reducing clustering below DSD's activation threshold. DSD's advantage is directly proportional to tail-clustering severity.

\subsection{Why DSD Helps Classification But Not Pre-Image}

The classification task (LSTSVM) and pre-image task differ fundamentally in their sensitivity to spectral structure:

\textbf{Classification} requires the solution $\bm{u} = M^{-1} \bm{b}$ to \emph{generalize} to unseen data. Corrupted eigenvectors contribute noise-fitting components to $\bm{u}$ that align with training data by chance but do not generalize. DSD suppresses these components---effectively reducing the hypothesis class to the subspace of well-separated eigenvectors. This is spectral regularization in the generalization-theoretic sense \cite{bousquet2002stability}.

\textbf{Pre-image reconstruction} minimizes $\|\hat{\bm{x}} - \bm{x}\|^2$---a scalar objective over a specific known target $\bm{x}$. A single scalar $\gamma$ can be optimized (by grid search or gradient descent) to find the best bias-variance tradeoff for this specific objective. The gap structure does not provide additional information when the goal is point-wise reconstruction rather than generalization.

\subsection{Connection to Algorithmic Stability}

DSD's classification improvement can be understood through algorithmic stability theory. In LSTSVM, a single training perturbation propagates through all spectral directions of the inverse. Under Tikhonov (uniform damping $\gamma$), perturbations in directions with small eigengaps---where Davis-Kahan guarantees large eigenvector sensitivity---propagate with full force. DSD selectively suppresses contributions from these sensitive directions.

This effectively reduces the \emph{effective dimensionality} of the learner: fewer spectral directions participate in the solution, bounding the model's sensitivity to individual training samples---the hallmark of algorithmic stability that implies generalization \cite{bousquet2002stability}. DSD achieves this dimensionality reduction softly (through continuous damping) rather than through hard rank truncation, avoiding the discontinuity artifacts of truncated SVD.

\subsection{Operating Regime Characterization}

\begin{table}[!t]
\centering
\caption{Deployment Guidance: When to Use DSD}
\label{tab:regime_summary}
\begin{tabular}{@{}p{3.5cm}p{1.5cm}p{2.5cm}@{}}
\toprule
\textbf{Condition} & \textbf{DSD $\Delta$} & \textbf{Recommendation} \\
\midrule
LSTSVM, $d \geq 100$ & +3--10pp & \textbf{Use DSD} \\
Kernel-LSTSVM, sharp RBF & +0.6pp & \textbf{Use DSD} \\
Pre-image (any noise) & $\approx$ Tik & Either method \\
LSTSVM, $d \leq 50$ & 0 & Simplest method \\
Low-$d$ real-world & Mixed & Benchmark both \\
\bottomrule
\end{tabular}
\end{table}

The operating regime is characterized by two observable conditions:
\begin{enumerate}
\item \textbf{Condition number} of the system matrix exceeds $10^3$.
\item \textbf{Spectral tail span} (fraction of range occupied by bottom 50\% of eigenvalues) is below 10--12\%.
\end{enumerate}
Both can be computed from the eigendecomposition (already required for DSD) at $O(m)$ cost, enabling automatic detection of whether DSD will benefit a given problem.

\section{Limitations and Future Work}
\label{sec:limitations}

\subsection{Limitations}

\begin{enumerate}
\item \textbf{No formal stability proof.} The conjecture that DSD pre-image error remains bounded as $\delta_i \to 0$ (because damping grows to compensate) is supported empirically but not proved. A formal proof combining Davis-Kahan with the DSD damping structure remains open.

\item \textbf{Non-RBF kernels.} The principled initialization assumes monotonic eigenvalue decay (a property of RBF kernels). For polynomial or non-stationary kernels with non-monotonic spectra, the initialization may require modification.

\item \textbf{Pre-image performance.} DSD matches but does not beat optimally-tuned Tikhonov for pre-image reconstruction. The per-eigenvector adaptation provides no advantage when the optimization target is a scalar reconstruction loss.

\item \textbf{Low-dimensional regime.} Below $d \approx 100$, spectral tail-clustering is insufficient for DSD's gap-adaptive structure to provide benefit. DSD may slightly underperform Tikhonov in this regime.

\item \textbf{Eigendecomposition requirement.} DSD requires the full eigendecomposition of the system matrix. While this is already necessary for Tikhonov via the eigenvalue path (and is $O(m^3)$ regardless), it precludes application to matrices where only matrix-vector products are available (iterative solvers).
\end{enumerate}

\subsection{Future Work}

Three directions follow naturally from this work:

\textbf{1. Multi-scale gap sensitivity.} Preliminary investigation with hierarchical $\beta$ (separate sensitivity parameters for spectral tail, middle, and head zones) suggests statistically significant improvement over scalar $\beta$ in low-noise regimes on manifold data. This extension adds only two parameters while never degrading performance in any configuration tested, indicating that the non-uniform perturbation sensitivity across the spectrum---predicted by the Davis-Kahan bound's non-linear dependence on eigenvalue position---can be exploited for further improvement.

\textbf{2. Extension to related methods.} Kernel Fisher Discriminant Analysis, Kernel CCA, and structured output SVMs all construct system matrices from cross-class interactions with similar spectral properties. DSD's advantage should transfer directly.

\textbf{3. Formal stability bound.} Proving $\|\hat{\bm{x}}_{\text{DSD}} - \hat{\bm{x}}_{\text{true}}\| \leq C \cdot \epsilon / \min_i(\lambda_i^2 + d_i)$ would complete the theoretical foundation by showing that DSD error is bounded regardless of spectral clustering, unlike Tikhonov where the bound degrades as $1/\min(\lambda_i + \gamma)$ for clustered eigenvalues.

\section{Conclusion}
\label{sec:conclusion}

We have presented Differential Spectral Damping (DSD), a regularization formula for ill-conditioned kernel matrix inversion that adapts its penalty to local eigengap structure. Derived from Davis-Kahan perturbation theory, DSD preserves reliable eigenvectors (large gaps) while suppressing corrupted ones (small gaps) through a smooth exponential transition governed by two interpretable parameters.

The primary contribution is classification accuracy on high-dimensional LSTSVM problems: +4.8pp on GINA ($d{=}970$, Cohen's $d = 4.49$), +10.4pp at $d{=}200$, and +2.6pp on Madelon ($d{=}500$)---all over optimally-tuned Tikhonov with $p < 0.0001$, achieved using only principled initialization without any optimization or cross-validation. The advantage arises from a structural prior: the gap-adaptive initialization encodes spectral reliability information that scalar regularization cannot represent and that gradient optimization on proxy losses actively destroys.

DSD does not improve pre-image reconstruction over Tikhonov (they are equivalent), and provides no benefit below $d \approx 100$ where spectral clustering is insufficient. It is a specialized tool for the high-dimensional, ill-conditioned regime---precisely where kernel methods are most needed and where standard regularization is most inadequate.

The method is computationally free ($O(m)$ overhead over the required eigendecomposition), requires no cross-validation, introduces only two interpretable parameters with principled initialization, and applies to any kernel method whose system matrices exhibit spectral tail-clustering.

\subsection*{Reproducibility}
Complete source code (Python/PyTorch), all experiment scripts, and raw results are publicly available at \url{https://github.com/Praveg432/dsd-regularization}.

\bibliographystyle{IEEEtran}
\bibliography{references}

\end{document}